\documentclass[10pt,twocolumn,letterpaper]{article}

\usepackage{cvpr}
\usepackage{times}
\usepackage{epsfig}
\usepackage{graphicx}
\usepackage{amsmath}
\usepackage{amssymb}

\usepackage{url}
\usepackage{subfigure}
\usepackage{amsmath}
\usepackage{amsthm}
\usepackage{amssymb}
\usepackage{algorithmic}
\usepackage{algorithm}
\usepackage{epstopdf}
\usepackage{cite}
\usepackage{multirow}
\usepackage{setspace}

\usepackage[pagebackref=False,breaklinks=true,letterpaper=true,colorlinks,bookmarks=false]{hyperref}

 \cvprfinalcopy 

\def\cvprPaperID{****} 

\ifcvprfinal\fi
\begin{document}

\pagenumbering{gobble}

\title{Deep Virtual Networks for Memory Efficient Inference of Multiple Tasks}

\author{Eunwoo Kim$^1$ \qquad Chanho Ahn$^2$ \qquad  Philip H.S. Torr$^1$ \qquad Songhwai Oh$^2$\\
$^1$University of Oxford \qquad $^2$Seoul National University\\
{\tt\small $\{$eunwoo.kim, philip.torr$\}$@eng.ox.ac.uk ~~  $\{$mychahn, songhwai$\}$@snu.ac.kr~~}
}

\maketitle

\begin{abstract}
Deep networks consume a large amount of memory by their nature.
A natural question arises can we reduce that memory requirement whilst maintaining performance.
In particular, in this work we address the problem of memory efficient learning for multiple tasks.
To this end, we propose a novel network architecture producing multiple networks of different configurations, termed deep virtual networks (DVNs), for different tasks.
Each DVN is specialized for a single task and structured hierarchically.
The hierarchical structure, which contains multiple levels of hierarchy corresponding to different numbers of parameters, enables multiple inference for different memory budgets.
The building block of a deep virtual network is based on a disjoint collection of parameters of a network, which we call a unit.
The lowest level of hierarchy in a deep virtual network is a unit, and higher levels of hierarchy contain lower levels' units and other additional units.
Given a budget on the number of parameters, a different level of a deep virtual network can be chosen to perform the task.
A unit can be shared by different DVNs, allowing multiple DVNs in a single network.
In addition, shared units provide assistance to the target task with additional knowledge learned from another tasks.
This cooperative configuration of DVNs makes it possible to handle different tasks in a memory-aware manner.
Our experiments show that the proposed method outperforms existing approaches for multiple tasks.
Notably, ours is more efficient than others as it allows memory-aware inference for all tasks.
\end{abstract}


\begin{figure}[t]
    \centering
    {\includegraphics[width=0.47\textwidth]{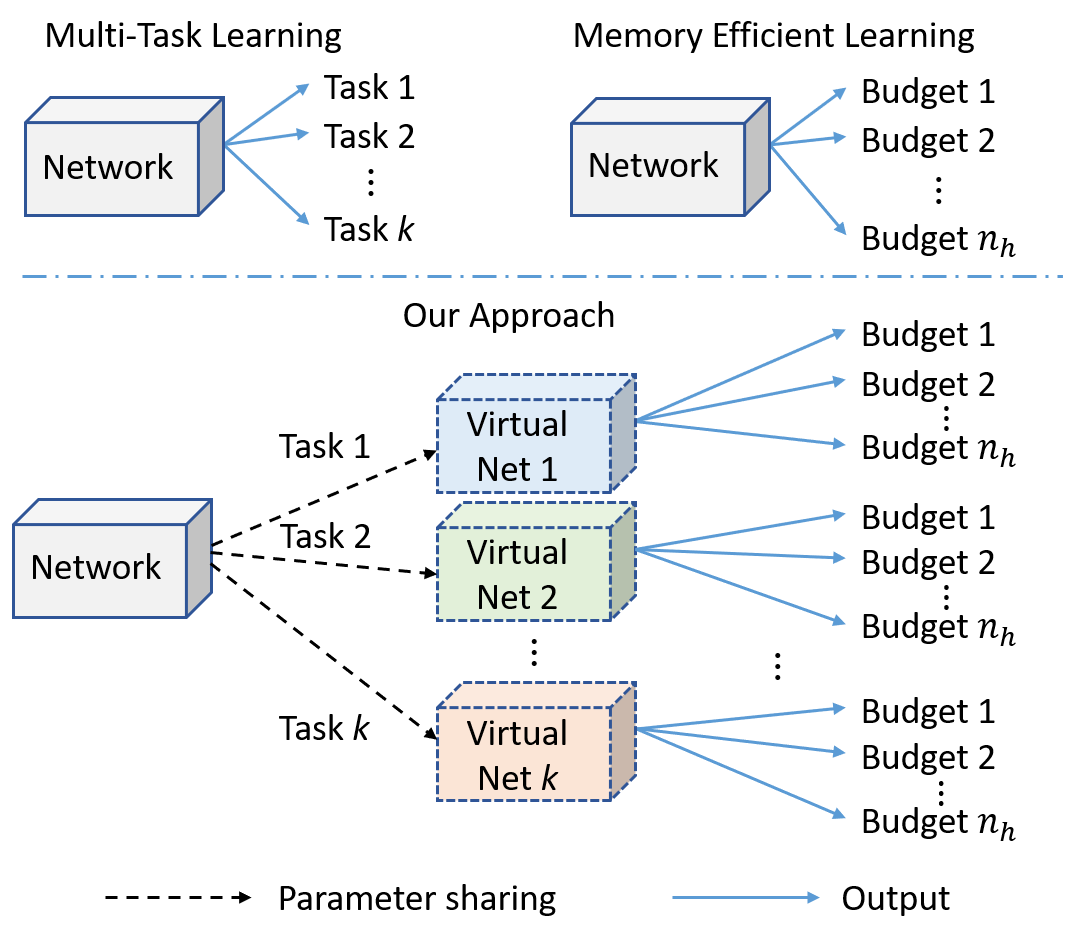}}
    \caption{
(Top left) Multi-task learning \cite{caruana1997multitask} for $k$
tasks and (top right) memory efficient learning \cite{kim2018nestednet} for a single task with $n_h$ different memory budgets, realized in a single network, respectively.
(Bottom) An overview of the proposed approach.
The proposed architecture contains multiple deep networks (deep
virtual networks) of different configurations for different tasks.
Each deep virtual network is specialized for a single task and allows
multiple inference for different memory budgets.
Our approach incorporates both multi-task and memory efficient
learning methods in a single architecture, producing $k \times n_h$
inference outputs, which significantly reduces the training efforts and
network storage.
    }
    \label{fig:overview}
\end{figure}

\section{Introduction}\label{sec:intro}
Recently, deep learning methods have made remarkable progress in computer vision and machine learning \cite{krizhevsky2012imagenet, simonyan2015very, he2016deep}.
Although successful in many applications,
it is well-known that many deep neural networks have a high memory footprint \cite{han2016eie, iandola2016squeezenet}.
This limits their practical applications, such as mobile phones, robots, and autonomous vehicles of low capacity.
The issue has been addressed by research aimed at reducing the number of parameters of a deep network to create a lightweight network \cite{han2015learning, he2017channel}.

Unfortunately, developing such a compact network is accompanied by a tradeoff between accuracy and the number of parameters (referred as the memory\footnote{We call the number of parameters as memory throughout the paper.}) at test time \cite{han2016deep, howard2017mobilenets}.
This requires efforts to find a proper network that gives competitive performance under a given memory budget \cite{gordon2018morphnet}.
Besides, when a network model with a different memory budget is required, we define and train a new network, which incurs additional training cost.

Recently, several studies have been conducted on multiple inference under different memory budgets in a single trained architecture \cite{larsson2017fractalnet, kim2018nestednet}, called memory efficient inference.
This problem can be achieved by designing a network structure (e.g., nested \cite{kim2018nestednet} and fractal \cite{larsson2017fractalnet} structures) which enables multiple inference corresponding to different memory budgets.
It allows flexible accuracy-memory tradeoffs within a single network and thus can avoid introducing multiple networks for different memory budget.
Note that memory budget may vary when tasks are performed simultaneously in a memory-limited device  (e.g., an autonomous vehicle with real-time visual and non-visual inference tasks to process at once).

Obviously, memory efficient inference can be an efficient strategy to provide different predictions in a network.
However, prior works have applied the strategy to a single task learning problem individually \cite{veit2017convolutional, kim2018nestednet}, and addressing multiple tasks jointly (often called multi-task learning \cite{caruana1997multitask, ruder2017overview}) with the strategy has been considered less.
Learning multiple tasks\footnote{Multiple tasks refer to multiple datasets, unless stated otherwise.} simultaneously in a network can have a single training stage and reduce the number of networks \cite{caruana1997multitask, mallya2018packnet}.
This approach also has the potential to improve generalization performance by sharing knowledge that represents associated tasks \cite{collobert2008unified, zhang2014facial, girshick2015fast}.
Despite its compelling benefits, little progress has been made so far in connection with memory efficient inference.
This is probably due to the difficulty of constructing a single network that allows memory efficient inference for different tasks.
The difficulty lies in the structural limitation of a neural network to possess a different structure for each task.

\begin{figure}[t]
    \centering
    \includegraphics[width=0.51\textwidth]{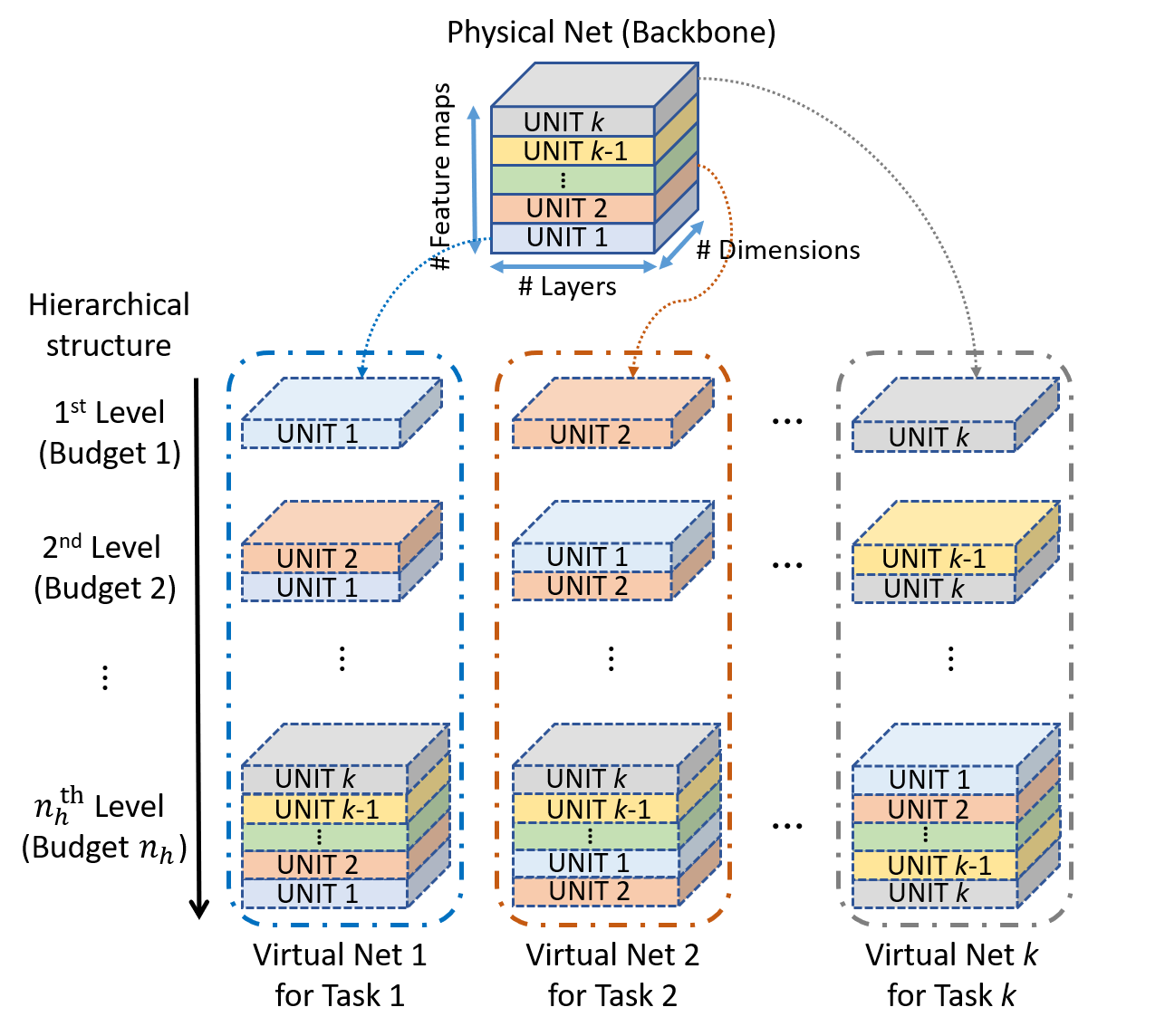}
    \caption{
A graphical illustration of the proposed approach which is based on a
backbone architecture (Physical Net) with $k$ preassigned disjoint structures, called units.
For a simple illustration, we assume that the number of feature maps
and their dimensions are the same across all layers (here, we omit fully-connected layers).
The proposed architecture produces $k$ deep virtual
networks (Virtual Nets), sharing its units for $k$ tasks.
A deep virtual network has a unique hierarchical structure with a
different order of units and is specialized for a designated task.
The number of levels of hierarchy in a deep virtual network is $n_h$, which corresponds to the number of different memory budgets.
This allows $k \times n_h$ inference for $k$ deep virtual networks.
(Best viewed in color.)
    }
    \label{fig:coop_rep}
\end{figure}

In this work, we aim to develop an efficient deep learning approach that performs memory efficient inference for multiple tasks in a single network.
To this end, we propose a novel architecture containing multiple networks of different configurations termed \textit{deep virtual networks (DVNs)}.
Each DVN shares parameters of the architecture and performs memory efficient inference for its corresponding task.
A virtual network resembles a virtual machine \cite{popek1974formal} in a computer system as multiple virtual machines can share resources of a physical computer.
Figure \ref{fig:overview} gives an overview of the proposed approach.

The proposed architecture is based on a backbone architecture, and we divide the network parameters into multiple disjoint sets along with their corresponding structures termed units.
Specifically, units are collected by dividing a set of feature maps in each layer into multiple subsets throughout the layers in the architecture (see Figure \ref{fig:coop_rep}).
A DVN is structured hierarchically which contains multiple levels of hierarchy corresponding to different numbers of units,
and a lower level of hierarchy assigns fewer units and a higher level of hierarchy contains more units.
For example, the lowest level of the hierarchy has a single unit.
Each level of the hierarchy in a DVN contains all preceding lower levels' units and one additional unit.
Hence, different levels of hierarchy in a DVN enables multiple inference
according to different memory budgets.
In the proposed architecture, a unit can be shared by different DVNs.
This allows multiple DVNs in a single deep network
for multiple tasks.
Each deep virtual network has a unique configuration (i.e., a hierarchical structure with a different order of units), and is specialized for a single task.
The unique configuration is determined by the proposed rule discussed in Section \ref{sec:main-ours}.
The proposed approach can selectively provide an inference output from its DVNs for a given task with the desired memory budget.
The approach is realized in a single training stage based on a single backbone architecture (e.g., a residual network \cite{he2016deep}), which significantly reduces training efforts and network storage.

We apply our method to joint learning scenarios of multiple tasks using popular image classification datasets.
Our results show that for all tasks DVNs are learned successfully under different memory budgets.
Even more, the results are better than other approaches. 
We also measure the actual processing time during inference to verify the practicality of the proposal.
In addition, we demonstrate our approach on the task of sequential
learning \cite{li2016learning}.

The proposed approach introduces a new concept of virtual networks in deep learning to perform multiple tasks in a single architecture, making it highly efficient.

\section{Related Work}\label{sec:related_work}

\noindent \textbf{Multi-task learning.} The aim of multi-task learning \cite{caruana1997multitask} is to improve the performance of multiple tasks by jointly learning them.
Two popular approaches are learning a single shared architecture with multiple output branches \cite{long2017learning, li2016learning} and learning multiple different networks according to tasks \cite{misra2016cross, yang2016trace}.
We are particularly interested in multi-task learning with a single shared network as it is memory efficient.
Recently, a few approaches have been proposed to perform multiple tasks in a single network by exploiting unnecessary redundancy of the network \cite{mallya2018packnet, kim2018nestednet}.
PackNet \cite{mallya2018packnet} divides a set of network parameters into multiple disjoint subsets to perform multiple tasks by iteratively pruning and packing the parameters.
NestedNet \cite{kim2018nestednet} is a collection of networks of different sizes which are constructed in a network-in-network style manner.
However, for a fixed budget the size of the assigned parameters of each network will be reduced as the number of tasks increases, which may cause a decrease in performance.
Moreover, they can produce an inference output for each task.
Whereas, our approach can overcome the issues by introducing deep virtual networks sharing disjoint subsets of parameters in our architecture and their different configurations make it possible to address multiple tasks (see Figure \ref{fig:coop_rep}).

Multi-task learning can be extended to sequential learning \cite{li2016learning, zenke2017continual, chaudhry2018riemannian}, where tasks are learned sequentially without accessing the datasets of old tasks.
Following the popular strategy in \cite{li2016learning}, we apply the proposed approach to sequential learning problems (see Section \ref{sec:main-ours-cont} and \ref{sec:exp-cont}).
\vspace{0.2cm}

\noindent \textbf{Memory efficient learning.} Memory efficient learning is a learning strategy to perform multiple inference according to different budgets on the number of parameters (called memory) in a single network \cite{larsson2017fractalnet, zamir2017feedback, kim2018nestednet}.
It enables flexible inference under varying memory budget, which is often called the anytime prediction \cite{zilberstein1996using}.
To realize the anytime prediction, a self-similarity based fractal structure \cite{larsson2017fractalnet} was proposed.
A feedback system based on a recurrent neural network \cite{zamir2017feedback} was proposed to perform different predictions according to memory or time budgets.
A nested network \cite{kim2018nestednet},
which consists of multiple networks of different scales, was proposed to address different memory budget.
However, these approaches are confined to performing an individual task.
In contrast, our method enables the anytime prediction for
multiple tasks using deep virtual networks.

To our knowledge, this work is the first to introduce deep virtual
networks of different configurations from a single deep
network, which enables flexible prediction under varying memory
conditions for multiple tasks.

\section{Approach}\label{sec:main}
\subsection{Memory efficient learning}\label{sec:main-pre}
We discuss the problem of memory efficient learning to perform multiple inference with respect to different memory budgets for a single task.
Assume that given a backbone network we divide the network parameters into $k$ disjoint subsets, i.e., $\mathcal{W} = [W_1, W_2, ..., W_k]$.
We design the network to be structured hierarchically by assigning the subsets, in a way that the $l$-th level of hierarchy ($l \geq 2$) contains the subsets in the ($l-1$)-th level and one additional subset \cite{kim2018nestednet}.
The lowest level of the hierarchy ($l=1$) assigns a single subset and the highest level contains all subsets (i.e., $\mathcal{W})$.
For example, when $k=3$ we can assign $W_1$ to the lowest level in the hierarchy, $[W_1, W_2]$ to the intermediate level, and $[W_1, W_2, W_3]$ to the highest level.
A hierarchical structure is determined by an order of subsets, which is designed by a user before learning.
In this work, the number of levels of hierarchy, denoted as $n_h$, is set to the number of subsets, $k$.
Each level of hierarchy defines a network corresponding to the subsets and produces an output.
The hierarchical structure thus enables $n_h$ inference for $n_h$ different numbers of subsets (memory budgets).

Given a dataset $\mathcal{D}$ consisting of image-label pairs and $n_h$ levels of hierarchy ,
the set of parameters $\mathcal{W}$
can be optimized by the sum of $n_h$ loss functions
\begin{equation}\label{eq:nested}
\min_{\mathcal{W}} ~  \sum_{l=1}^{n_h} \mathcal{L}\Big(h^l (\mathcal{W}); \mathcal{D}\Big),
\end{equation}
where $h^l(\mathcal{W})$ is a set of parameters of $\mathcal{W}$ that
are assigned to the $l$-th level of hierarchy.
There is a constraint on $h$ such that a higher level set includes a lower level set, i.e., $h^p(\mathcal{W}) \subseteq h^q(\mathcal{W}), p \leq q, \forall p,q \in [1,...,n_h]$, for a structure sharing parameters \cite{kim2018nestednet}.
$\mathcal{L}(\cdot)$ is a standard loss function (e.g., cross-entropy) of a network associated with $\mathcal{D}$.
In addition, we enforce regularization on $\mathcal{W}$ (e.g., $l_2$ decay) for improved learning.
By solving (\ref{eq:nested}), a learned network is collected and can perform $n_h$ inference corresponding to $n_h$ memory budgets.

\begin{figure*}[t]
    \centering
    {\includegraphics[width=0.89\textwidth]{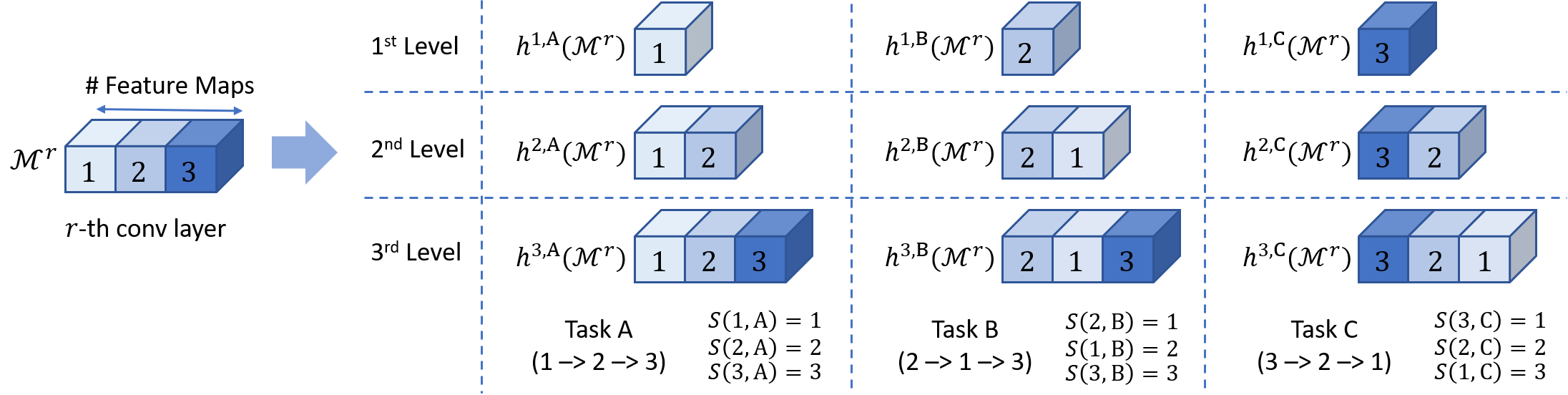}}
    \caption{
An example of constructing three different hierarchical structures in
the $r$-th convolutional layer of a network, denoted as
$\mathcal{M}^r$, which consists of three disjoint sets of feature maps
or units (i.e., $\mathcal{M}^r = [{M}^r_1, {M}^r_2, {M}^r_3]$ and the
number indicates the unit index).
The number of tasks and the number of levels of hierarchy are three.
Different orders of the units construct the hierarchical structures.
Here, $h^{l, j}(\mathcal{M}^r)$ is a function that selects the sub-structure of
$\mathcal{M}^r$ corresponding to the $l$-th level of hierarchy for the $j$-th task.
$S(i,j)$ returns the level number at which the $i$-th unit ${M}^r_i$
  is added to the hierarchy for the $j$-th task.
(Best viewed in color.)
    }
    \label{fig:conv_group}
\end{figure*}

The function $h^l(\mathcal{W})$ can be designed by a pruning operation on $\mathcal{W}$ in element-wise \cite{han2015learning}
or group-wise (for feature maps) \cite{li2017pruning}.
Since our approach targets a practical time-dependent inference,
we follow the philosophy of group-wise pruning approaches \cite{wen2016learning, he2017channel} in this work.
Note that the problem (\ref{eq:nested}) is applied to a single task (here, a dataset $\mathcal{D}$), rarely considering multiple tasks (or datasets).
This issue will be addressed in the following subsection with the introduction of deep virtual networks.

\subsection{Deep virtual network}\label{sec:main-ours}
\noindent \textbf{Building block.} Our network architecture is based on a backbone architecture, and we divide the network parameters into multiple disjoint subsets.
Assume that there are $k$ disjoint subsets in a network, which are collected by dividing feature maps in each layer into $k$ subsets across all layers.\footnote{For simplicity, we omit a fully-connected layer. However, it is appended on top of the last convolutional layer to produce an output.}
Formally, a set of network parameters is represented as $\mathcal{W} = \{\mathcal{W}^r\}_{1 \leq r \leq L}$, where $L$ is the number of layers and $\mathcal{W}^r = [W_1^r, W_2^r, ..., W_k^r] \in \mathbb{R}^{w^r \times h^r \times c_I^r \times c_o^r}$.
The $i$-th subset of $\mathcal{W}^r$ is denoted as $W_i^r \in \mathbb{R}^{w^r \times h^r \times c_I^r(i) \times c_o^r(i)}$.
Here, $w^r$ and $h^r$ are the width and height of the convolution kernel of the $r$-th layer, respectively.
$c_I^r$ and $c_o^r$ are the number of input and output feature maps of the $r$-th layer, respectively, such that $\sum_{j=1}^k c_I^r(j) = c_I^r$ and $\sum_{j=1}^k c_o^r(j) = c_o^r$.
The set of the $i$-th subsets over all layers is written as
\begin{equation}\label{eq:weight_rep}
W_i = [W_i^1, W_i^2, ..., W_i^{L}].
\end{equation}
We call the corresponding network structure defined by $W_i$ as \textit{unit $i$},
which produces an inference output.
\vspace{0.15cm}

\noindent \textbf{Hierarchical structure.} The proposed approach produces deep virtual networks (DVNs) of different network configurations (i.e., hierarchical structures) using shared units in a network architecture, as illustrated in Figure \ref{fig:coop_rep}.
Each unit is coupled with other units along the feature map direction to form a hierarchical structure similar to the strategy described in Section \ref{sec:main-pre}.
The number of levels of hierarchy is $n_h$, where a level of the hierarchy includes all preceding lower levels' units and one additional unit.
A different hierarchical structure is constructed by a different order of units.
This introduces a unique DVN which is specialized for a task.
Thus, multiple DVNs of different network configurations can be realized in a
single network by sharing units, for different tasks (see Figure \ref{fig:conv_group}).
Whereas, the problem (\ref{eq:nested}) is for a single task with a network configuration, which is equivalent to producing a single DVN.
\vspace{0.15cm}

\noindent \textbf{Rules for configuring virtual networks.} In order to
determine different network configurations of deep virtual networks,
we introduce a simple rule.
We assume that datasets are collected sequentially, along with their task ID numbers, and the datasets with adjacent task ID numbers are from similar domains.
The proposed rule is:
(i) The unit $i$ is assigned to the task $i$, and it becomes the lowest level in the hierarchy for the task.
(ii) The unit $i$ is coupled with adjacent units that are not coupled.
(iii) If there are two adjacent units, the unit with a lower task ID number is coupled.
For example, assume that $h^{l,j}(\mathcal{W})$ is a function that selects the subset of $\mathcal{W}$ of the $l$-th level of hierarchy for the task $j$.
When $k=3$ and $\mathcal{W} = [W_1, W_2, W_3]$,
where $W_i$ denotes the parameters for the unit $i$,
we construct the following hierarchical structure\footnote{When units are used together, additional parameters (interconnection between units) are added to parameters of stand-alone units, $W_i$'s.} from the rule for the task $j$
\begin{equation}
\begin{split}
h^{1,j}(\mathcal{W}) = &W_j, \qquad \qquad ~~~~~~~~~~~~~~~~~~~~~~\mbox{if} ~~ 1 \leq j \leq k, \\
h^{2,j}(\mathcal{W}) =
&\begin{cases}
~ [W_1, W_{2}], \qquad \qquad ~~~~~~~~\mbox{if} ~~ j=1, \\
~ [W_{j}, W_{j-1}], \qquad \quad ~~~~~~~~\mbox{if} ~~ 1<j\leq k, \\
\end{cases}\\
h^{3,j}(\mathcal{W}) =
&\begin{cases}
~ [W_1, W_2, W_3], \qquad \qquad \mbox{if} ~~ j=1 , \\
~ [W_{j}, W_{j-1}, W_{j+1}], \qquad \mbox{if} ~~ 1 < j < k, \\
~ [W_{j}, W_{j-1}, W_{j-2}], \qquad \mbox{if} ~~ j=k.\\
\end{cases}\\
\end{split}
\raisetag{50pt}
\end{equation}
The configuration is different depending on the order of units (see Figure \ref{fig:conv_group} for an example).
\vspace{0.15cm}

\noindent \textbf{Objective function.}
Given datasets for $k$ tasks, $\mathcal{T} = [\mathcal{D}^1, \mathcal{D}^2, ..., \mathcal{D}^{k}]$,
$k$ deep virtual networks,
the set of parameters $\mathcal{W}$,
and $n_h$ levels of hierarchy for each deep virtual network,
the proposed method can be optimized by solving the sum of $k \times n_h$ loss functions
\begin{equation}
\min_{\mathcal{W}} ~ \mathcal{L}^k(\mathcal{W}) \triangleq
\sum_{j=1}^{k} \left(\sum_{l=1}^{n_h} \mathcal{L}\Big(h^{l,j}(\mathcal{W}); \mathcal{D}^j\Big) \right),
\label{eq:resource-eff}
\end{equation}
where $h^{l,j}(\mathcal{W})$ is a function that selects the subset of
$\mathcal{W}$ corresponding to the $l$-th level of hierarchy for the
$j$-th task (or deep virtual network $j$),
such that $h^{p,j}(\mathcal{W}) \subseteq h^{q,j}(\mathcal{W}),  p \leq q, \forall p, q \in [1, ..., n_h]$, for all $j \in [1,...,k]$.
Note that in the case when $k=1$, the problem (\ref{eq:resource-eff}) reduces to the problem (\ref{eq:nested}) for a single task \cite{kim2018nestednet}.
\vspace{0.15cm}

\noindent \textbf{Learning.} The unit $i$ is learned based on the following gradient with respect to $W_i$
\begin{equation}\label{eq:coopnet_grad}
\frac{\partial \mathcal{L}^k(\mathcal{W})}{\partial W_i} =
\sum_{j=1}^k \left( \sum_{l = S(i,j)}^{n_h} \frac{\partial \mathcal{L}_{l,j}(\mathcal{W})}{\partial W_i} \right),
\end{equation}
where $\mathcal{L}_{l,j}(\mathcal{W}) \triangleq \mathcal{L}(h^{l,j}(\mathcal{W}); \mathcal{D}^j)$.
$S(i,j)$ returns the level number at which the $i$-th unit is added to the hierarchy for the $j$-th task (see Figure \ref{fig:conv_group}).
The unit $i$ is learned by aggregating multiple gradients from the
hierarchical structures of deep virtual networks for all $k$ tasks.
Note that, for given $n_h$, the difference $|n_h - S(i,j)|$ influences
on the amount of the gradient (significance) of the unit $i$ for the
task $j$ as the gradients from more levels accumulate.
As the difference is larger, the significance of the unit will be higher for the task $j$.
The proposed approach is trained in a way that each unit is learned to
have different significance (different $S(i,j)$) for all tasks.
Note that the total amount of gradients of a unit over all tasks is
about same to those of other units using the proposed configuration rule.
This prevents units from having irregular scales of gradients.

\subsection{Deep virtual network for sequential tasks}\label{sec:main-ours-cont}

The proposed approach can also handle sequential tasks \cite{li2016learning}.
Assume that the old tasks, from the first to the $(k-1)$-th
task, have been learned beforehand.
For the current (new) task $k$, we construct an architecture with $k$ units, where $k-1$ units correspond to the old tasks and the $k$-th unit represents the current task.
Based on the units, we construct $k$ deep virtual networks as described in Section \ref{sec:main-ours}.

Given a dataset for the task $k$, $\mathcal{D}^k$,
the set of parameters $\mathcal{W}$,
$k$ deep virtual networks,
and $n_h$ levels of hierarchy,
the problem ($k > 1$) is formulated as
\begin{equation}\label{eq:resource-eff2}
\min_{\mathcal{W}} \sum_{j=1}^{k-1} \left( \sum_{l=1}^{n_h} \mathcal{L}_D^j\Big(h^{l,j}(\mathcal{W}); \mathcal{D}^k\Big)\right)
+ \sum_{l=1}^{n_h}  \mathcal{L}\Big(h^{l,k}(\mathcal{W}); \mathcal{D}^k\Big),
\end{equation}
where $\mathcal{L}_D^j(h^{l,j}(\mathcal{W)}; \mathcal{D}^k)$ is a distillation loss between the output of a network whose corresponding structure is determined by $h^{l,j}(\mathcal{W})$ and the output of the task $j$ from the old network when a new input $\mathcal{D}^k$ is given.
The only exception from the problem (\ref{eq:resource-eff}) (which jointly learns $k$ tasks) is that we use a distillation loss function $\mathcal{L}_D^j(\cdot)$ to preserve the knowledge of the old tasks in the current sequence \cite{li2016learning} (due to the absence of the old datasets).
For $\mathcal{L}_D^j(\cdot)$, we adopt the modified cross entropy function \cite{hinton2015distilling} following the practice in \cite{li2016learning}.
The gradient of (\ref{eq:resource-eff2}) with respect to $W_i$ is
\begin{equation}
\sum_{j=1}^{k-1} \left( \sum_{l = S(i,j)}^{n_h} \frac{\partial
  \mathcal{L}_{D(l,j)}^j(\mathcal{W})}{\partial W_i} \right) + \sum_{l
  = S(i,k)}^{n_h} \frac{\partial \mathcal{L}_{(l,k)}(\mathcal{W})}{\partial W_i},
\label{fig:coop_grad}
\end{equation}
where $\mathcal{L}_{D(l,j)}^j (\mathcal{W})$ $\triangleq$ $\mathcal{L}_D^j( h^{l,j}(\mathcal{W}); \mathcal{D}^k)$ and $\mathcal{L}_{(l,k)}(\mathcal{W}) \triangleq \mathcal{L}(h^{l,k}(\mathcal{W}); \mathcal{D}^k)$.
In the special case where $n_h = 1$ and $h^{l,j}(\mathcal{W}) = \mathcal{W}$, $\forall l,j$, the problem (\ref{eq:resource-eff2}) reduces to the learning without forgetting (LwF) problem \cite{li2016learning}, which has the following gradient
\begin{equation}
\sum_{j=1}^{k-1} \frac{\partial \mathcal{L}_{D}^j(\mathcal{W}; \mathcal{D})}{\partial \mathcal{W}} + \frac{\partial \mathcal{L}(\mathcal{W}; \mathcal{D})}{\partial \mathcal{W}}.
\end{equation}
Compared to our gradient in (\ref{fig:coop_grad}), LwF learns a single set of parameters $\mathcal{W}$, which reveals that the network has no hierarchical structure and all tasks are performed without memory efficient inference.

\section{Experiments}\label{sec:exp}
\subsection{Experimental setup}\label{sec:exp-setup}
We tested our approach on several supervised learning problems using visual images.
The proposed method was applied to standard multi-task learning (joint learning) \cite{caruana1997multitask}, where we learn multiple tasks jointly, and sequential learning \cite{li2016learning}, where we focus on the $k$-th sequence with the learned network for old tasks.
We also applied the proposed approach to hierarchical classification \cite{yan2015hd}, which is the problem of classifying coarse-to-fine class categories.
Our approach was performed based on four benchmark datasets:
CIFAR-10 and CIFAR-100 \cite{krizhevsky2009learning},
STL-10 \cite{stl-10}, 
and Tiny-ImageNet\footnote{\url{https://tiny-imagenet.herokuapp.com/}}, 
based on two popular (backbone) models, WRN-$n$-$s$ \cite{zagoruyko2016wide} and ResNet-$n$ \cite{he2016deep}, where $n$ and $s$ are the number of layers and the scale factor over the number of feature maps, respectively.

We first organized three scenarios for joint learning of multiple tasks.
We performed a scenario (\textit{J1}) consisting of two tasks using the CIFAR-10 and CIFAR-100 datasets and another scenario (\textit{J2}) of four tasks whose datasets are collected by dividing the number of classes of Tiny-ImageNet into four subsets evenly.
The third scenario (\textit{J3}) consists of three datasets,
CIFAR-100, Tiny-ImageNet, and STL-10, of different image scales (from 32$\times$32 to 96$\times$96).
For hierarchical classification (\textit{H1}), CIFAR-100 was used
which contains coarse classes (20 classes) and fine classes (100 classes).
For sequential learning, we considered two scenarios,
where a scenario (\textit{S1}) has two tasks whose datasets are collected by dividing the number of classes of CIFAR-10 into two subsets evenly,
and another scenario (\textit{S2}) consists of two tasks using CIFAR-10 and CIFAR-100.

We compared the proposed approach with other recent approaches handling multiple tasks: Feature Extraction \cite{donahue2014decaf},
LwF \cite{li2016learning},
DA-CNN \cite{wang2017growing},
PackNet \cite{mallya2018packnet}, and
NestedNet \cite{kim2018nestednet}.
We also compared with the backbone networks, ResNet \cite{he2016deep} and WRN \cite{zagoruyko2016wide}, as baseline approaches performing an individual task.

\subsection{Implementation details}\label{sec:exp-impl}
All the compared architectures were based on ResNet \cite{he2016deep} or WRN \cite{zagoruyko2016wide}.
We followed the practice of constructing the number of feature maps in residual blocks in \cite{he2016deep} for all applied scenarios.
We constructed the building block of a network for Tiny-ImageNet based on the practice of ImageNet \cite{he2016deep}.
All the compared methods were learned from scratch until the same epoch number and were initialized using the Xavier method \cite{glorot2010understanding}.
The proposed network was trained by the SGD optimizer with Nesterov momentum of 0.9, where the mini-batch sizes were 128 for CIFAR and 64 for Tiny-ImageNet, respectively.
We adopted batch normalization \cite{ioffe2015batch} after each convolution operation.

We constructed units with respect to feature maps across the convolution layers, except the first input layer.
Our deep virtual networks have task-specific input layers for different tasks or input scales, respectively, and the dimensionality of their outputs are set to the same by varying the stride size using convolution.
When two units are used together, the feature map size doubles and additional parameters (i.e., interconnection between the units) are needed to cover the increased feature map size, in addition to parameters (intraconnection) of stand-alone units.
We also appended a fully connected layer of a compatible size on top of each level of hierarchy.
All the proposed approaches were implemented under the TensorFlow library \cite{abadi2016tensorflow}, and their evaluations were provided based on an NVIDIA TITAN Xp graphics card.

\subsection{Joint learning}\label{sec:exp-joint}
We conducted experiments for joint learning
by comparing with two approaches: PackNet$^+$  (a grouped variant of PackNet \cite{mallya2018packnet}  to achieve actual inference speed-up by dividing feature maps into multiple subsets similar to ours), and NestedNet (with channel pruning) \cite{kim2018nestednet} which can perform either multi-task learning or memory efficient learning.

\begin{figure}[t]
    \centering
    {\includegraphics[width=\columnwidth]{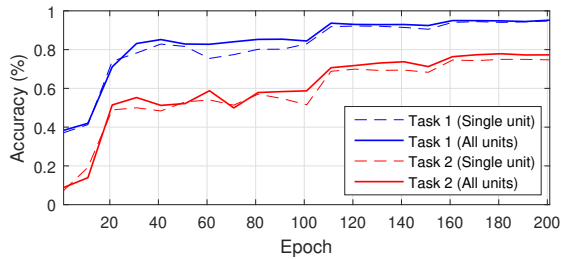}}
    \caption{
    Performance curve of the proposed DVNs for the joint learning on CIFAR-10 (task 1) and CIFAR-100 (task 2).}
    \label{fig:joint_cifar_curve}
\end{figure}

For the first scenario (\textit{J1}) using the two CIFAR datasets,
we split the number of parameters almost evenly along the feature map
dimension and assigned the first half and all of the parameters to the
first and second task, respectively, for PackNet$^+$ and NestedNet.
Our architecture contains two deep virtual networks (DVNs), and each DVN consists of two units (and two levels of hierarchy) by splitting a set of feature maps in every layer into two subsets evenly throughout all associated layers.
Here, each stand-alone unit has 25$\%$ of the parameter density, since
inter-connected parameters between the two units are ignored
(see Section \ref{sec:exp-impl}).
For this scenario, WRN-32-4 \cite{zagoruyko2016wide} was used for all compared approaches.
Table \ref{tab:joint_cifar} shows the results of the compared approaches.
Our approach gives four evaluations according to tasks and memory budgets.
Among them, the evaluations using each stand-alone unit (top) do not compromise much on performance compared to those using all units (bottom) on average.
PackNet$^+$ and NestedNet give the comparable performance to our approach, but their maximum performance leveraging the whole network capacity are poorer than ours.
Baseline gives comparable performance to the multi-task learning approaches, but it requires 2$\times$ larger number of parameters in this problem.
The average inference times (and the numbers of parameters) of our DVN using single and all associated units
are 0.11ms (1.9M) and 0.3ms (7.4M) for a single image, respectively.
We also provide the performance curve of the proposed approach on the test sets in Figure \ref{fig:joint_cifar_curve}.

\begin{table}[t]
\caption{Results of joint learning on CIFAR-10 (task 1) and CIFAR-100 (task 2).
$N_O$ and $N_T$ are the number of inference outputs produced by a method and the total number of parameters of a method, respectively.
Baseline results are collected from two independent networks.
Ours provides two different inference using a single unit (top) and all units (bottom) for each task.
\strut
}
\centering
    \renewcommand\arraystretch{0.95}
    \small
    \begin{tabular}{ c | c | c | c | c | c } \hline
     Method & $N_O$ & $N_T$ & Task 1 & Task 2 & Average \\ \hline \hline
     Baseline \cite{zagoruyko2016wide} & 1 & 14.8M & 94.8$\%$ & 76.4$\%$ & 85.6$\%$  \\
     PackNet$^+$ {\footnotesize \cite{mallya2018packnet}} & 2 & 7.4M & 94.5$\%$ & 75.3$\%$ & 84.9$\%$  \\
     NestedNet {\footnotesize\cite{kim2018nestednet}}     & 2 & 7.4M & 94.7$\%$ & 76.7$\%$ & 85.7$\%$ \\  \hline
     \multirow{2}{*}{Ours} & \multirow{2}{*}{4} & \multirow{2}{*}{7.4M}  & 94.6$\%$ & 75.0$\%$ & 84.8$\%$ \\
                           &                    &  & 95.1$\%$ & 77.3$\%$ & 86.2$\%$ \\ \hline
    \end{tabular}
    \label{tab:joint_cifar}
\end{table}

\begin{figure*}[t]
    \centering
    \subfigure[Multi-task learning]
    {\includegraphics[width=1.0\columnwidth]{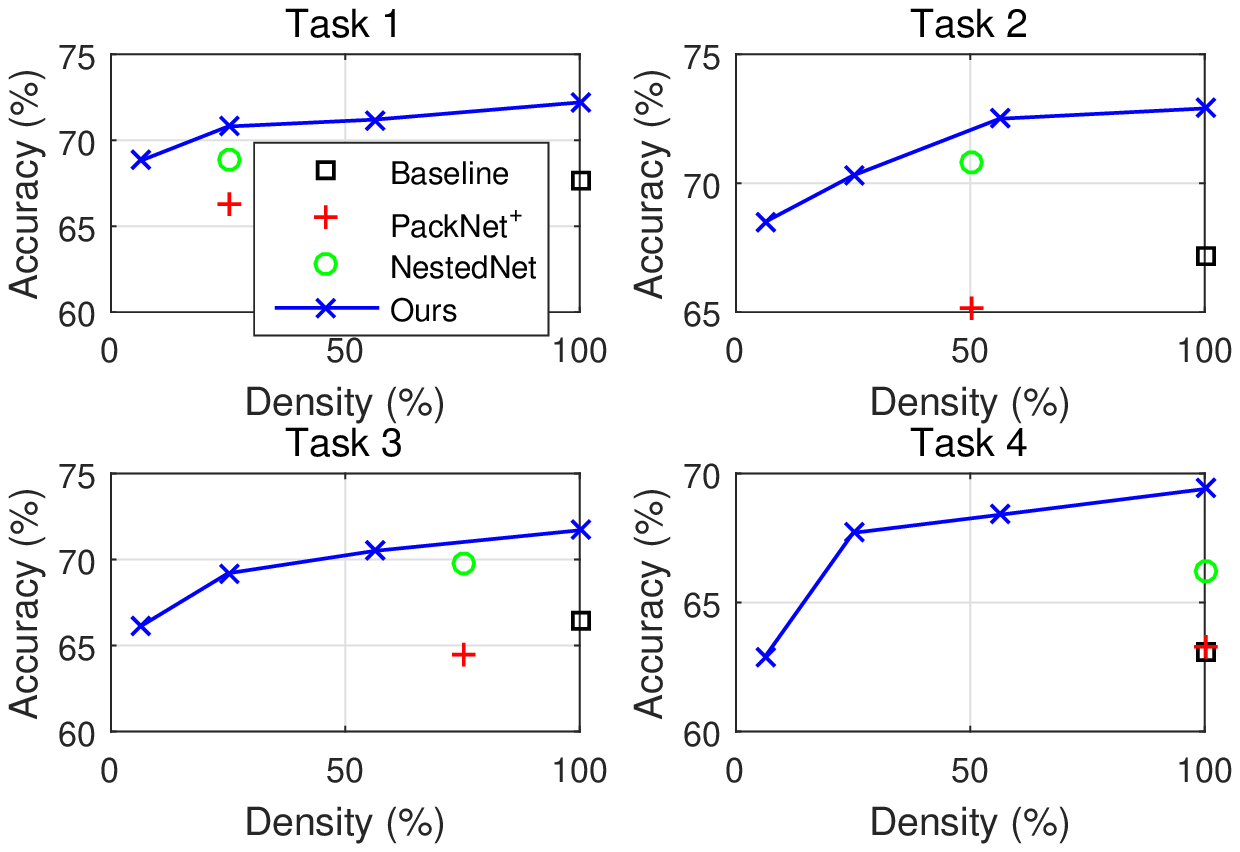}  
    \label{fig:joint_tiny}}
    \subfigure[Memory efficient learning]
    {\includegraphics[width=0.935\columnwidth]{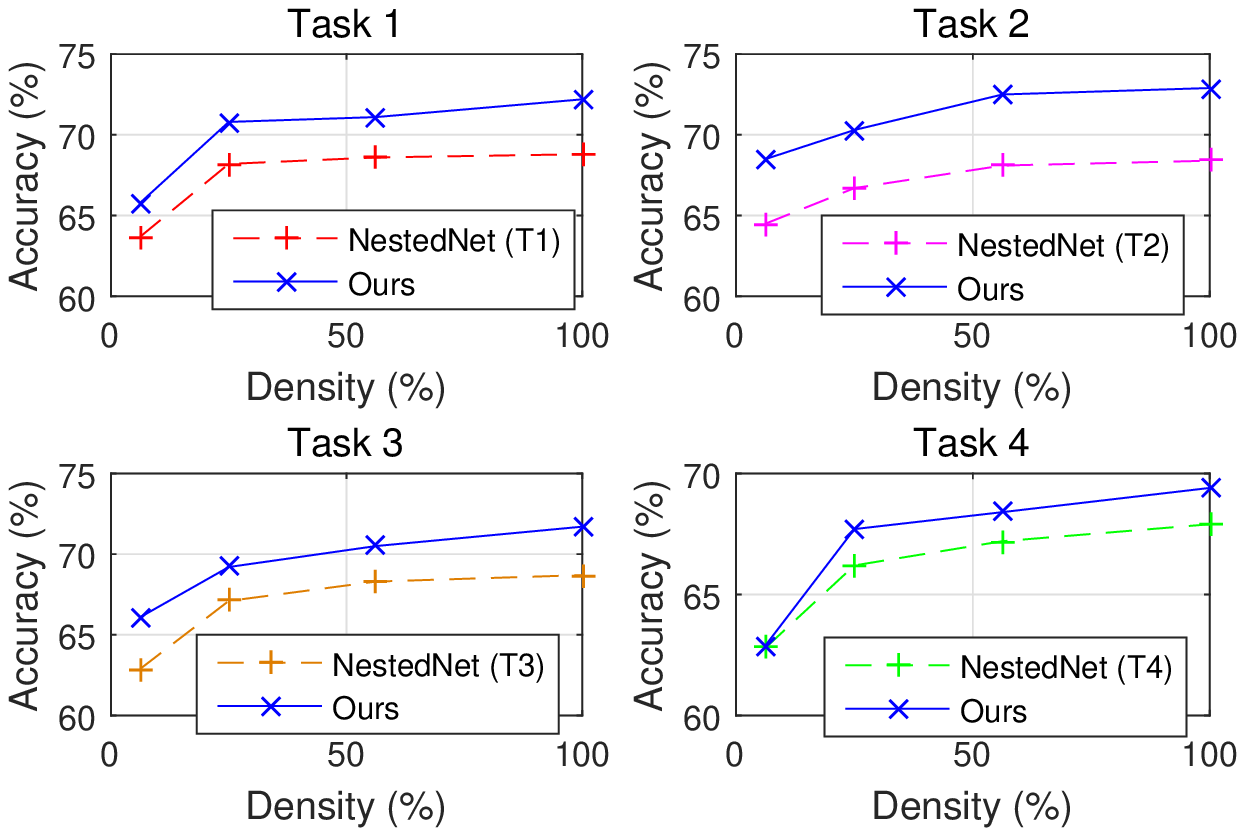} 
    \label{fig:joint_tiny_multiway}}
    \caption{
    Results of joint learning on the Tiny-ImageNet tasks with respect to parameter density ratios (budgets).
    (a) Multi-task learning:
    Each deep virtual network in our approach provides four evaluations with respect to different parameter density ratios for each task, while other methods produce an evaluation with a fixed budget.
    Baseline requires four trained networks to achieve the results.
    (b) Memory efficient learning:
    Our deep virtual networks produce $4 \times 4$ inference outputs within a single trained network, while NestedNet requires four different trained networks to perform memory efficient inference for the same tasks, respectively.
    ($\cdot$) denotes the task ID.
    }
    \label{fig:cont_cifar}
\end{figure*}

Figure \ref{fig:joint_tiny} shows the results for the second scenario (\textit{J2}) using Tiny-ImageNet (four tasks).
The ratios of parameters for PackNet$^+$ and NestedNet were
$\frac{1}{4}:\frac{2}{4}:\frac{3}{4}:1$ from task 1 to task 4, by
dividing parameters into four subsets almost evenly and assigning
the first $j$ subsets to task $j$.
Our architecture contains four DVNs each of which has four units and four levels of hierarchy.
The ratios of parameters in each hierarchy were $\frac{1}{16}:\frac{4}{16}:\frac{9}{16}:1$ for each DVN.
All compared approaches were based on ResNet-42 \cite{he2016deep}.
As shown in the figure, our approach outperforms the competitors under similar memory budgets for all tasks.
Moreover, ours provides additional outputs for different memory budgets, making it highly efficient.
Even though NestedNet has the similar strategy of sharing parameters, it performs poorer than ours.
Unlike the previous example, the baseline shows unsatisfying results and even requires $4 \times$ larger network storage than ours to perform the same tasks.

\begin{table}[t]
\caption{Parameter density and speedup of our method with respect to levels of hierarchy.
$l(i)$ denotes the level containing $i$ units.
\strut
}
\centering
    \renewcommand\arraystretch{0.95}
    \small
    \begin{tabular}{ c | c | c | c | c } \hline
                          & $l(1)$   & $l(2)$    & $l(3)$    & $l(4)$  \\ \hline \hline
    No. parameters        & 1.9M     &  7.5M     & 16.8M     & 29.8M \\
    Density               & 6.4$\%$  & 25.2$\%$  & 56.4$\%$  & 100$\%$  \\
    Compression rate      & 15.7$\times$ & 4.0$\times$  & 1.8$\times$  & 1$\times$ \\ \hline
    Inference time (ms)   & 0.18         &  0.38        & 0.67         & 1.05 \\
    Practical speed-up    & 5.8$\times$  & 2.8$\times$  & 1.6$\times$  & 1$\times$ \\  \hline
    \end{tabular}
    \label{tab:params_analysis}
\end{table}

In addition, we compared with NestedNet \cite{kim2018nestednet} on the same scenario (\textit{J2}) for memory efficient inference.
Since NestedNet performs memory efficient inference for a task,
we trained it four times according to the number of tasks.
Whereas, our architecture was trained once and performed memory efficient inference for all the tasks from our DVNs.
Figure \ref{fig:joint_tiny_multiway} shows that our method gained significant performance improvement over NestedNet for all the tasks.
Table \ref{tab:params_analysis} summarizes the number of parameters and its associated speed-ups of the proposed network.

\begin{figure}[t]
    \hspace{-7mm}
    \includegraphics[width=1.15\columnwidth]{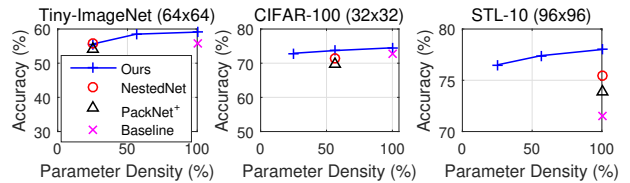}
    \caption{
    Results on three different datasets (Tiny-ImageNet, CIFAR-100, and STL-10) of different scales for joint learning.
    }
    \label{fig:three_sets}
\end{figure}

For the third scenario (\textit{J3}) on three different tasks, a set of feature maps is divided into three subsets for the compared methods.
The ratios of parameters were $\frac{4}{16}: \frac{9}{16}: 1$ from
task 1 (Tiny-ImageNet) to task 3 (STL-10).
Each DVN has the same density ratios in its hierarchical structure.
ResNet-42 \cite{he2016deep} was applied by carefully following the
network design and learning strategy designed for ImageNet
\cite{he2016deep}.
Figure \ref{fig:three_sets} shows the results for the tasks.
The proposed method performs better than the compared approaches on average under similar parameter density ratios.
While PackNet$^+$ and NestedNet give comparable performance to ours for
Tiny-ImageNet, they perform poorer than ours for the other two tasks.
Moreover, they produce a single output for every task with a fixed parameter density condition, while ours provides multiple outputs under different density conditions for each dataset.
The numbers of parameters and their inference times of our DVN are 0.65ms (7.5M), 1.02ms (16.8M), and 1.51ms (29.8M), respectively, for a single image from STL-10.

\begin{table}[t]
\caption{Results of the hierarchical classification on CIFAR-100.
$(\cdot)$ denotes the number of classes.
$N_O$ is the number of inference outputs produced by a method.
Baseline results are collected from independent networks.
NestedNet provides two different results according to the number of tasks.
Our approach performs four different inference, according to the number of parameters and tasks.\strut}
\centering
    \renewcommand\arraystretch{1.0}
    \small
    \begin{tabular}{ c | c | c | c | c | c } \hline
      & $N_O$ & \multicolumn{2}{c|}{Task 1 (20)}   & \multicolumn{2}{|c}{Task 2 (100)} \\ \hline
     No. parameters                    &  $-$ &1.8M & 7.4M & 1.8M & 7.4M   \\ \hline \hline
     Baseline \cite{zagoruyko2016wide} & 1  & 82.1$\%$  & 84.9$\%$  & 73.4$\%$  & 75.7$\%$    \\
 	 NestedNet \cite{kim2018nestednet} & 2 & 83.7$\%$   & $-$       & $-$       & 76.6$\%$    \\
  	 Ours                              & 4  & 84.1$\%$  & 86.1$\%$  & 74.9$\%$  & 76.9$\%$    \\ \hline
    \end{tabular}
    \label{tab:joint_hc}
\end{table}

\begin{table*}[t]
\caption{Results of the sequential learning on the CIFAR-10 tasks.
The proposed architecture contains two deep virtual networks each of which provides two different evaluations using a single unit (right column) and all the units (left column) for each task.
}
\centering
    \renewcommand\arraystretch{1.0}
    \small
    \begin{tabular}{ c | c | c | c | c | c | c } \hline
     Method & Feature Extraction \cite{donahue2014decaf} & DA-CNN \cite{wang2017growing} & LwF \cite{li2016learning} & NestedNet \cite{kim2018nestednet} & \multicolumn{2}{c}{Ours}  \\ \hline \hline
     Task 1   & 96.3$\%$  & 96.3$\%$  & 95.3$\%$  & 93.9$\%$  & 95.4$\%$  & 95.8$\%$ \\
	 Task 2   & 85.7$\%$  & 90.1$\%$  & 97.1$\%$  & 98.2$\%$  & 97.7$\%$  & 98.1$\%$ \\ \hline
     Average  & 91.0$\%$  & 93.2$\%$  & 96.2$\%$  & 96.05$\%$ & 96.55$\%$ & 96.95$\%$ \\ \hline
    \end{tabular}
    \label{tab:cont_cifar10}
\end{table*}

\begin{table*}[t]
\caption{Results of the sequential learning on the CIFAR-10 (task 1) and CIFAR-100 (task 2) datasets.
}
\centering
    \renewcommand\arraystretch{1.0}
    \small
    \begin{tabular}{ c | c | c | c | c | c | c } \hline
     Method & Feature Extraction \cite{donahue2014decaf} & DA-CNN \cite{wang2017growing} & LwF \cite{li2016learning} & NestedNet \cite{kim2018nestednet} & \multicolumn{2}{c}{Ours}  \\ \hline \hline
     Task 1  & 94.9$\%$  & 94.9$\%$  & 93.4$\%$  & 93.1$\%$  & 93.1$\%$  & 93.4$\%$  \\
	 Task 2  & 53.2$\%$  & 57.4$\%$  & 77.2$\%$  & 77.9$\%$  & 78.0$\%$  & 78.7$\%$  \\ \hline
     Average & 74.05$\%$ & 76.15$\%$ & 85.3$\%$  & 85.5$\%$  & 85.55$\%$ & 86.05$\%$ \\ \hline
    \end{tabular}
    \label{tab:cont_tiny}
\end{table*}

\subsection{Hierarchical classification}\label{sec:exp-hc}
As another application of joint learning, we experimented with the scenario (\textit{H1}), hierarchical classification \cite{yan2015hd}.
The aim is to model multiple levels of hierarchy of class category for a dataset, and each level is considered as a task.
We evaluated on CIFAR-100 which has two-level hierarchy
of class category as described in Section \ref{sec:exp-setup}.
Our architecture contains two deep virtual networks, and each contains two units by dividing feature maps equally into two sets.
Thus, it produces four different inference outputs.
We compared with NestedNet \cite{kim2018nestednet} which can perform hierarchical classification in a single network.
The backbone network was WRN-32-4.

Table \ref{tab:joint_hc} shows the results of the applied methods.
We also provide the baseline results by learning an individual network (WRN-32-2 or WRN-32-4) for the number of parameters and the number of classes.
Overall, our approach performs better than other compared methods for all cases.
Ours and NestedNet outperform the baseline probably due to their property of sharing parameters between the tasks as they are closely related to each other.
The proposed approach produces a larger number of inference outputs than NestedNet while keeping better performance.

\subsection{Sequential learning}\label{sec:exp-cont}
We conducted the scenario (\textit{S1}) which consists of two sequential tasks based on CIFAR-10, where the old (task 1) and new (task 2) tasks consist of the samples from the first and last five classes of the dataset, respectively.
We compared our approach with other methods that can perform sequential tasks: Feature Extraction \cite{donahue2014decaf}, LwF \cite{li2016learning}, DA-CNN \cite{wang2017growing} (with two additional fully-connected layers), and NestedNet \cite{kim2018nestednet} (whose low- and high-level of hierarchy in the network represent old and new tasks, respectively).

The proposed network consists of two units by dividing feature maps into two subsets evenly (each stand-alone unit has 25$\%$ parameter density ratio).
It constructs two deep virtual networks providing four inference outputs.
We applied the WRN-32-4 architecture for all compared approaches.
Table \ref{tab:cont_cifar10} shows the results of the compared methods.
We observe that the proposed approach outperforms other approaches.
Notably, the results using stand-alone units are better than others on average.
Feature Extraction and DA-CNN nearly preserve the performance for the first task by maintaining the parameters of the first task unchanged, but their performances give the unsatisfactory results for the following task.
Whereas, the results from LwF and NestedNet are much better than those mentioned above for the second task, but their results are worse than ours.

We also applied the proposal to another scenario (\textit{S2}) consisting of CIFAR-10 (old, task 1) and CIFAR-100 (new, task 2).
All the compared approaches were performed based on WRN-32-8.
Our DVNs were constructed and trained under the same strategy to (\textit{S1}).
The results of the scenario are summarized in Table \ref{tab:cont_tiny}.
Our result using all units (right column) gives the best performance on average among the compared approaches.
Moreover, our result using a stand-alone unit (left column) also performs better than the best competitors, LwF and NestedNet, which use the same distillation loss function \cite{hinton2015distilling}.

\section{Conclusion}\label{sec:conclusion}

In this work, we have presented a novel architecture producing deep virtual networks (DVNs) to address multiple objectives with respect to different tasks and memory budgets.
Each DVN has a unique hierarchical structure for a task and enables multiple inference for different memory budgets.
Based on the proposed network, we can adaptively choose a DVN and one of its level of hierarchy for a given task with the desired memory budget.
The efficacy of the proposed method has been demonstrated under various multi-task learning scenarios.
To the best of our knowledge, this is the first work introducing the concept of virtual networks in deep learning for multi-task learning.

\vspace{4mm}
{\small
\noindent \textbf{Acknowledgements.}
This work was supported by the ERC grant ERC-2012-AdG 321162-HELIOS, EPSRC grant Seebibyte EP/M013774/1, EPSRC/MURI grant EP/N019474/1, Basic Science Research Program
through the National Research Foundation of Korea (NRF) funded by the
Ministry of Science and ICT (NRF-2017R1A2B2006136), and
AIR Lab (AI Research Lab) of Hyundai Motor Company through HMC-SNU
AI Consortium Fund.
We would also like to acknowledge the Royal Academy of Engineering and FiveAI.
}

{\small
\bibliographystyle{ieee_fullname}
\bibliography{egbib}
}

\end{document}